\documentclass[10pt,twocolumn,letterpaper]{article}

\usepackage[pagenumbers]{cvpr} 
\usepackage{soul} %
\usepackage{color, xcolor} %

\definecolor{cvprblue}{rgb}{0.21,0.49,0.74}
\usepackage[pagebackref,breaklinks,colorlinks,citecolor=cvprblue]{hyperref}

\usepackage{cite}
\usepackage{times}
\usepackage{epsfig}
\usepackage{graphicx}
\usepackage{amsmath}
\usepackage{amssymb}
\usepackage{comment}
\usepackage{multirow}
\usepackage{bm}
\usepackage[ruled]{algorithm2e}

\def\paperID{1592} 
\def\confName{CVPR}
\def\confYear{2025}

\title{Revisiting Marr in Face: The Building of 2D--2.5D--3D Representations in \\ Deep Neural Networks}

\author{
Xiangyu Zhu$^{1,2}$, Chang Yu$^{1,3}$\thanks{Corresponding author}, Jiankuo Zhao$^{1,2}$, Zhaoxiang Zhang$^{1,2}$, Stan Z.Li$^{3}$, Zhen Lei$^{1,2,4}$\\
$^{1}$State Key Laboratory of Multimodal Artificial Intelligence Systems, \\
Institute of Automation, Chinese Academy of Sciences\\
$^{2}$School of Artificial Intelligence, University of Chinese Academy of Sciences\\
$^{3}$AI Lab, Research Center for Industries of the Future, Westlake University\\
$^{4}$Centre for Artificial Intelligence and Robotics, Hong Kong Institute of Science \& Innovation,\\
Chinese Academy of Sciences\\
{\tt\small \{xiangyu.zhu,zhaojiankuo2024,zhaoxiang.zhang,zhen.lei\}@ia.ac.cn}\\
{\tt\small \{yuchang, Stan.ZQ.Li\}@westlake.edu.cn}\\
}

\begin{document}
\maketitle
\hyphenpenalty=5000
\tolerance=1000

\begin{abstract}
David Marr's seminal theory of vision proposes that the human visual system operates through a sequence of three stages, known as the 2D sketch, the 2.5D sketch, and the 3D model. In recent years, Deep Neural Networks (DNN) have been widely thought to have reached a level comparable to human vision. However, the mechanisms by which DNNs accomplish this and whether they adhere to Marr's 2D--2.5D--3D construction theory remain unexplored.
In this paper, we delve into the face perception task to explore these questions and find evidence supporting Marr's theory. We introduce a graphics probe, a sub-network crafted to reconstruct the original face image from the network's intermediate layers. The key to the graphics probe is its flexible architecture that supports image reconstruction in both 2D and 3D formats, as well as in a transitional state between them.
By injecting graphics probes into neural networks, and analyzing their behavior in reconstructing images, we find that DNNs initially encode images as 2D representations in low-level layers, and finally construct 3D representations in high-level layers. Intriguingly, in mid-level layers, DNNs exhibit a hybrid state, building a geometric representation that captures surface normals within a narrow depth range, akin to the appearance of a low-relief sculpture. This stage resembles the 2.5D representations, providing a view of how DNNs evolve from 2D to 3D in the perception process. The graphics probe therefore serves as a tool for peering into the mechanisms of DNN, providing empirical support for Marr's theory.

\end{abstract}

\section{Introduction}\label{sec1}
\begin{figure}
   \begin{center}
   \includegraphics[width=0.98\linewidth]{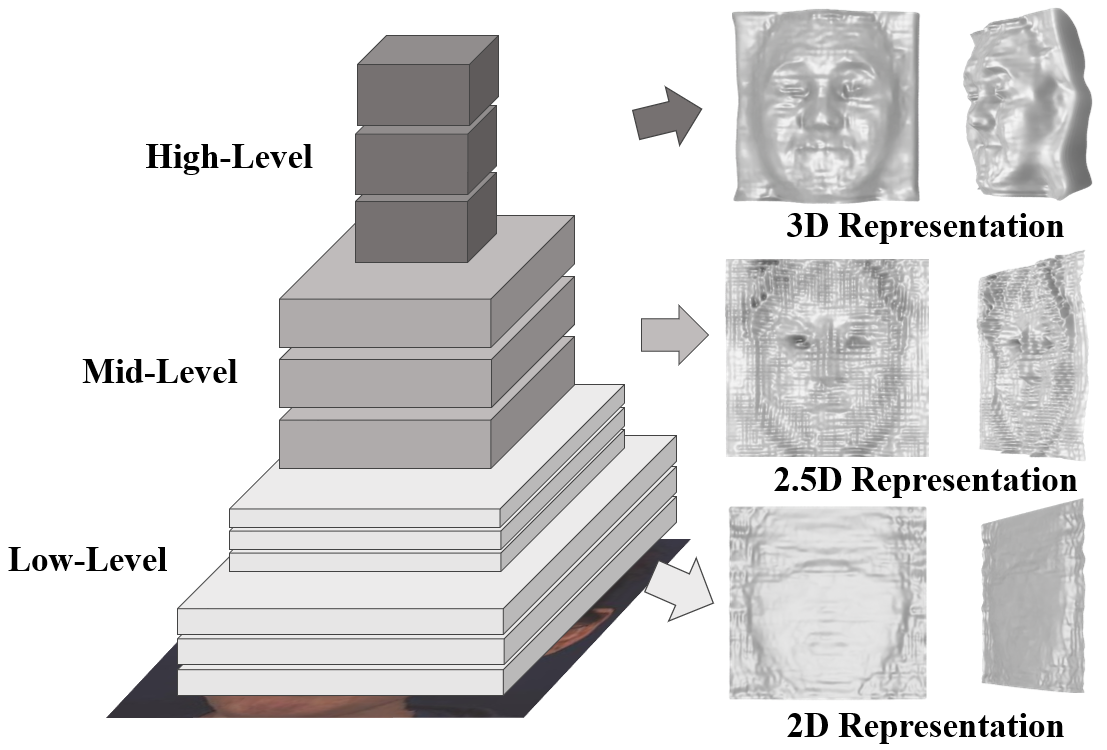}
   \caption{The building of 2D--2.5D--3D representations in DNN.}
   \label{fig-briefview}
   \end{center}
\end{figure}
One of the hallmarks of human face perception is its ability to extract robust object representations. This allows us to rapidly recognize a face under different situations. However, specific mechanisms underlying this skill remain unidentified, despite decades of research in psychology and neuroscience~\cite{marr1978representation, logothetis1995shape, poggio2004generalization,diwadkar1997viewpoint,yamins2016using, bansal2016marr,yildirim2020efficient,tacchetti2018invariant}. There has been a longstanding debate about whether objects are represented in an object-centered or viewer-centered manner~\cite{logothetis1995shape}. Object-centered representations are typically encoded with features that are independent of the viewpoint, often relying on 3D reconstructions~\cite{kazhdan2003rotation, liebelt2008independent}. On the other hand, viewer-centered representations store a collection of viewpoint-dependent features, allowing for matching a given view to the closest stored representation~\cite{liu2020recognizing,calder2011oxford}.
In David Marr's pioneering vision theory~\cite{marr1978representation}, the concepts of view-centered and object-centered representations are integrated through a three-stage process. First, the visual system generates a 2D primal sketch that represents local features of the stimulus. Next, a 2.5D sketch is created, offering a viewer-centered representation of the observed surfaces, typically comprising a field of local surface orientations. Finally, the information from the 2.5D sketches is integrated to construct an object-centered 3D model that comprehensively captures the complete 3D structure of the perceived object. Although these assumptions are partly supported by neurophysiological research~\cite{yamins2016using,logothetis1994view} and computational feasibility~\cite{bansal2016marr,yildirim2020efficient,tacchetti2018invariant}, the underlying mechanisms remain unknown.

Recent developments in DNN achieve outstanding results in many face perception-related tasks~\cite{masi2018deep,wu2019facial,li2020deep}, often outperforming human experts. These achievements generate the excitement that perhaps the algorithms essential to high-level face perception would automatically emerge in DNN~\cite{hill2019deep,o2018face,parde2017face}. 
However, a known limitation of current network architectures is their black-box nature, which hinders accessibility to understanding the representations encoded within them. Consequently, interpreting the contents of intermediate features and understanding the process through which the desired output is generated is significantly challenging. In this paper, we aim to find intuitive and visually comprehensible evidence that reveals how visual representations are constructed. We are particularly interested in exploring whether the construction pipeline aligns with the 2D--2.5D--3D framework proposed by David Marr in the 1980s.~\cite{marr1978representation}. 
To this end, we introduce a novel probing paradigm by injecting a new type of graphics probe into the intermediate layers of DNNs. These probes are designed to gather information from a group of neurons and draw the encoded content in an interpretable manner. 
Specifically, there are four operations, 1) A probing feature interact with the neurons within a receptive field through self-attention~\cite{vaswani2017attention} to gather the information they encode. 2) The probing feature is split into $K$ graphics probes using $K$ learned templates. 3) Each graphics probe is tasked with generating an image patch through Computer Graphics (CG) elements, including \textit{depth map}, \textit{albedo map}, \textit{view direction}, and \textit{Phong lighting}. 4) The assembly of these patches is required to reconstruct the input image. The key to the probing paradigm is its flexibility in the ways of image reconstruction. For instance, the depth map can be flat or have a 3D structure, the view direction can be object-centered or viewer-centered, and the learned templates indicate the concepts that DNNs grasp. The choices made in these aspects would shed light on the preferred perceptual behavior of DNNs.

In the experiments, we showcase the behavior of graphics probes across different layers. First, we observe that the depth map of the graphics probe initially presents as a flat plane at the bottom. This plane is then etched to introduce rich normal variations in the middle, which leads to the production of shading effects. Ultimately, the depth map evolves into a 3D model at the top. This progression from 2D to 2.5D and finally 3D supports Marr's foundational assumption about the vision process. Next, we investigate the variations in the view directions of the probes. We observe that in lower layers, the probed views are maintained as the canonical view, which is perpendicular to the depth map. In the upper layers, these views are adjusted to align with the true viewpoint. This evolution indicates a shift from a viewer-centered to an object-centered perspective. Finally, we delve into the formulation of the templates. We observe that in the mid-level layers, the templates tend to rely on the viewpoint of the input face, such as left, frontal, and right views. In the top layer, the templates shift focus to semantic components such as the forehead, facial features, and jaw, which are independent of specific viewpoints. This suggests that DNNs initially establish view-based prototypes and subsequently form a part-whole hierarchy in the top layer. Our contributions thus provide insights into several key questions, including: ``What are the intermediate representations of DNNs like?'' ``Are the representations template-based or 3D-based?'' and `Under what circumstances is a 3D representation established?''.

\begin{figure*}
   \begin{center}
   \includegraphics[width=0.99\linewidth]{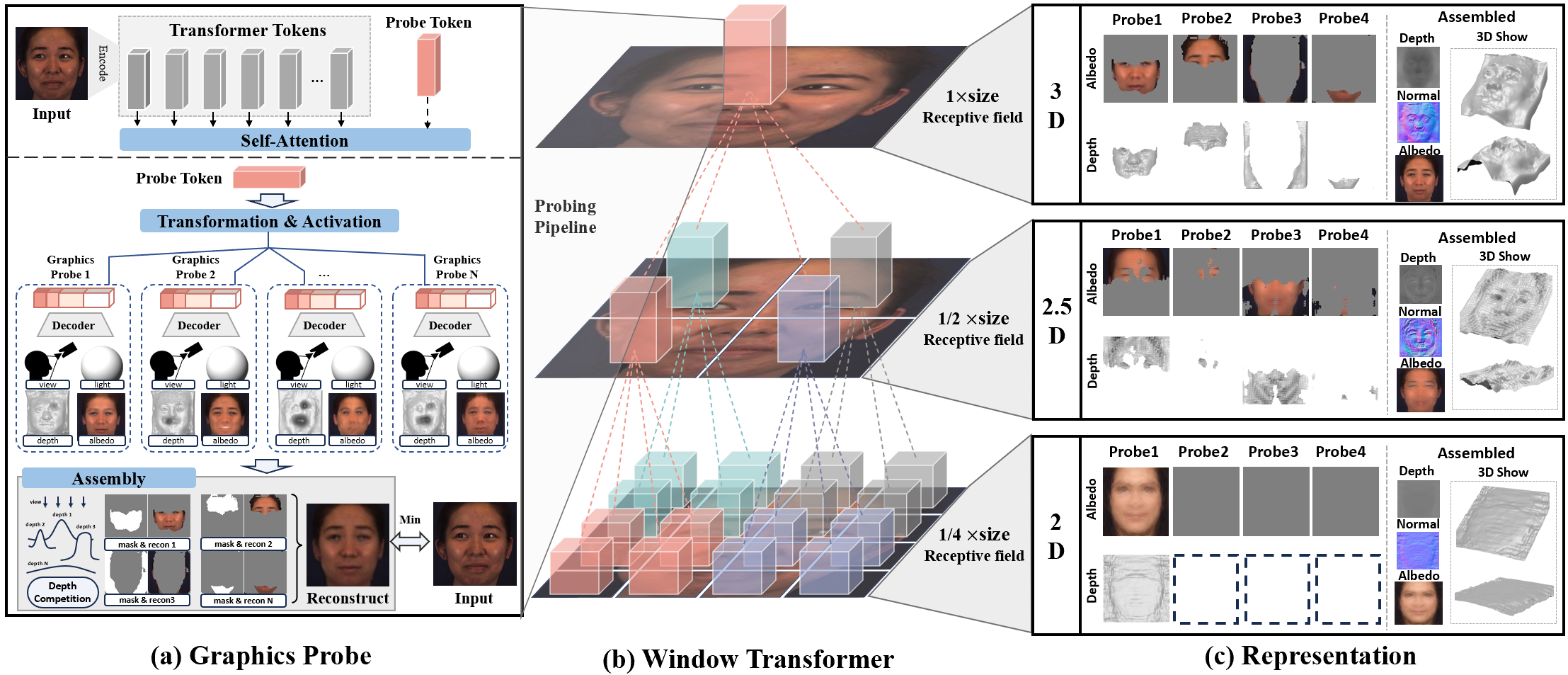}
   \caption{\textbf{Schematic of Graphics Probe}. (a) During the probing process, a probe token interacts with the original tokens and generates multiple graphics probes to reconstruct the input image in a CG manner. (b) The architecture of the probed network. (c) The visualization of probed representations across different levels: 2D at the low level, 2.5D at the middle level, and 3D at the high level.}
   \label{fig-method}
   \end{center}
\end{figure*}

\section{Related Work}

\subsection{Probing Intermediate Representations}
The probing technology~\cite{kim2018interpretability,alain2016understanding} employs a linear classifier to estimate specific concepts, thereby determining which concepts are captured in the internal layers of the network. In face recognition network, the probing results~\cite{zhong2016face,terhorst2021soft} reveal that intermediate representations contain rich information about non-identity attributes, including expressions and image conditions. Additionally, there is a rapid increase in identity expressivity in the top layer~\cite {dhar2020attributes}. An analysis of caricature face representations~\cite{hill2019deep}  reveals that variations in faces are organized in a hierarchical manner. Face identity is nested under gender, and illumination and viewpoint are nested under identity. IGC-Net~\cite{yu2023graphics} learns hierarchical 3D face representations by decomposing objects into semantically consistent part-level descriptions and then assembling them into object-level descriptions. Net2Vec~\cite{fong2018net2vec} aligns semantic concepts with filter activations and shows that a combination of filter responses is necessary to fully represent complex concepts. These findings may suggest that view-dependent responses at the middle level are associated together to build the view-independent representation at the top level. It is noteworthy that there is evidence suggesting an alignment between DNNs and human brains. Grossman et al.~\cite{grossman2019convergent} record neuronal activity in higher visual areas of human brains and find that face-selective responses exhibit similarity to those observed in intermediate layers of DNNs.

\subsection{3D Reconstruction from 2.5D sketch}
In single-image 3D reconstruction, an effective approach is to first extract a 2.5D sketch, such as a depth or normal map, before proceeding to reconstruct the full 3D geometry. This strategy is based on the observation that 2.5D sketches can be more easily extracted from 2D images and are more readily transferred from synthetic to real-world scenarios~\cite{wu2017marrnet}. Wu et al.~\cite{wu2017marrnet} and Sun et al.~\cite{sun2018pix3d} utilize an end-to-end model to sequentially predict 2.5D sketches and the corresponding 3D shape from RGB images. An alternative approach by Lun et al.~\cite{lun20173d} infers depth and normal maps from line drawings and subsequently refines them into 3D point clouds through energy minimization techniques. ShapeHD~\cite{wu2018learning} further employs an adversarially trained regularizer to discourage the generation of unrealistic shapes. As a further intermediate step between 2.5D sketches and full 3D reconstruction, GenRe~\cite{zhang2018learning} introduces spherical maps, allowing the completion of non-visible object surfaces based on the visible ones. 
Although these studies demonstrate that employing DNNs to generate a 2.5D sketch before 3D reconstruction can be advantageous, it remains an open question whether DNNs naturally develop a sequential 2D--2.5D--3D representation during image processing, especially in the absence of explicit 2.5D or 3D supervision.

\section{Method}
We investigate the mechanism of DNN by inserting the graphics probes into a modified window-transformer architecture. As shown in Figure~\ref{fig-method}(a), within each window, a probe token is inserted and interacts with the existing tokens via self-attention. This probe token is then transformed and activated into several graphics probes. Each graphical probe consists of several visually comprehensible components, including depth map, albedo map, view, and lighting. By rendering these graphic probes to reproduce the image, the intermediate latent features can be visualized in an interpretable CG manner. Our findings support the hypothesis that object representation is constructed through a 2D--2.5D--3D sequence, as depicted in Figure~\ref{fig-method}(c).

\subsection{Probed Architecture}
The probed architecture is a Window Transformer (WinT)~\cite{yu2022degenerate}, which is a modification of the Swin-Transformer~\cite{liu2021swin} without the window shifting mechanism. WinT utilizes a group of tokens to perceive a local image window through self-attention mechanisms~\cite{dosovitskiy2020image}. These tokens are iteratively aggregated~\cite{vaswani2017attention} to form higher-level tokens with an increasing window size, culminating in a group of tokens that encompasses the entire image, as shown in Figure~\ref{fig-method}(b). In WinT, tokens have clearly defined receptive fields. Besides, the encoded content can be effectively accessed by the injected probe token through self-attention, facilitating easy and controllable probing. In the experiments, we additionally tested various architectures to assess the generalizability of our findings.

WinT splits the feature map into non-overlapping windows and assigns $T$ tokens to each of them to encode its content, with self-attention being applied exclusively within each window. After several Transformer blocks, the number of tokens is reduced by concatenating the features of each group of $2 \times 2$ neighboring windows, leading to a $2 \times$ downsampling in resolution. Thus, the WinT process can be viewed as four hierarchical stages from bottom to up, corresponding to window sizes that are $\frac{1}{8}\times$, $\frac{1}{4}\times$, $\frac{1}{2}\times$, and $1 \times$ of the original image size.  For each window, a probe token is inserted to probe the content. Assuming that at the final layer of a particular stage, the image $\mathbf{I}$ is encoded into window-based tokens by the transformer network $\mathcal{E}$:
\begin{equation}
\begin{split}
    &\{\mathbf{E}_{1}, \mathbf{E}_2, \ldots 
    , \mathbf{E}_K\} = \mathcal{E}(\mathbf{I})\\
    &\mathbf{E}_k = \{\mathbf{e}^k_1, \mathbf{e}^k_2,  \ldots , \mathbf{e}^k_M, \mathbf{p}_{k}\}
\end{split}
\end{equation}
where $\mathbf{E}_{k}$ is a group of tokens that encode the visual information within a window, and $K$ is the number of windows, which is $\{64,16,4,1\}$ at each stage. Each group of tokens includes $M$ visual tokens $\mathbf{e}$ just as a traditional transformer, with an additional probe token $\mathbf{p}$ inserted. During encoding, self-attention is applied across $\mathbf{E}_{k}$, enabling the probe token to gather information. Subsequently, the probe tokens $\{\mathbf{p}_1, \mathbf{p}_2, \ldots, \mathbf{p}_K\}$ from all the windows are utilized to reconstruct the image.

\subsection{Graphics Probe}
In graphics probe, a probe token $\mathbf{p}$ is transformed to CG elements and rendered to an image. Specifically, for each layer, we maintain $K$ learned bases as the templates to generate $K$ graphics probes:
\begin{equation}\label{equ-template}
\begin{split}
    &\mathbf{W} = [\mathbf{w}_{1}, \mathbf{w}_{2}, ..., \mathbf{w}_{K}] \in \mathbb{R}^{D \times K}, \\
    &\mathbf{A} = [\mathbf{p}_{1} \odot \mathbf{w}_{1}, \mathbf{p}_{2} \odot \mathbf{w}_{2}, ..., \mathbf{p}_{k} \odot \mathbf{w}_{K}] \in \mathbb{R}^{D \times K},\\
    &\mathbf{A}_{[d,:]} = {\rm hardmax}(\mathbf{A}_{[d,:]}), ~~~ d=1,2, \ldots ,D, \\
    &\mathbf{\theta}_k = \mathbf{p}_k  \odot \mathbf{A}_{[:,k]},  ~~~ k=1,2, \ldots ,K,
\end{split}
\end{equation}
where each column of the matrix $\mathbf{W}$ serves as a template, $K$ is the number of basis and $D$ is the token dimension. We compute the dot product between the probe token $\mathbf{p}$ and its template to generate an attention matrix $\mathbf{A}$, whose $d$th row is $\mathbf{A}_{[d,:]}$ and $k$th column is $\mathbf{A}_{[:,k]}$. Subsequently, the hardmax function is applied to each row of $\mathbf{A}$, resulting in a one-hot encoded vector. Finally, the dot product is computed between the probe tokens and columns of $\mathbf{A}$, yielding $K$ graphics probes  $\{\mathbf{\theta}_1, \mathbf{\theta}_2, ..., \mathbf{\theta}_K\}$. Through this procedure, only the dimension with the highest activation with its template is left. It is noteworthy that there is only one probe token at the highest level, while we still maintain multiple templates and replicate the probe token to match these templates, thereby creating multiple graphics probes.

Each graphics probe $\mathbf{\theta}_{k}$ is the concatenation of four components: geometry $\mathbf{\theta}^g_{k}$, albedo $\mathbf{\theta}^a_{k}$, view $\mathbf{\theta}^v_{k}$, and lighting $\mathbf{\theta}^l_{k}$. Each component can be translated to a CG element by a corresponding decoder: 
\begin{equation}\label{equ-decompose}
\begin{split}
    \mathbf{\theta}_{k}&=\{\mathbf{\theta}_{k}^g, \mathbf{\theta}_{k}^a, \mathbf{\theta}_{k}^v, \mathbf{\theta}_{k}^v\},  \\
    \mathbf{D}_{k} &= \mathcal{D}_g(\mathbf{\theta}_{k}^g) \in \mathbb{R}^{64 \times 64}, \\
    \mathbf{A}_{k} &= \mathcal{D}_a(\mathbf{\theta}_{k}^b) \in \mathbb{R}^{64 \times 64 \times 3}, \\
    \mathbf{V}_{k} &= \mathcal{D}_v(\mathbf{\theta}_{k}^v) \in \mathbb{R}^{6},\\
    \mathbf{L}_{k} &= \mathcal{D}_l(\mathbf{\theta}_{k}^l) \in \mathbb{R}^{4},
\end{split}
\end{equation}
where $\mathcal{D}_g, \mathcal{D}_a, \mathcal{D}_v, \mathcal{D}_l$ are independent decoders used to transform features into a depth map $\mathbf{D}_{k}$, an RGB albedo map $\mathbf{A}_{k}$, a 6DoF camera view $\mathbf{V}_{k}$, and ambient/direct lighting $\mathbf{L}_{k}$, respectively, as shown in Figure~\ref{fig-method}(a).  
In the rendering process, not all graphics probes are employed to render the image. Instead, they are assembled through an depth competition process. At each pixel, only the graphics probe with the highest depth is rendered, akin to Z-buffer:
\begin{equation}\label{equ-assemble}
\begin{aligned}
\mathbf{M}_k(i,j)=\mathbf{1}&_{k=\mathop{\mathrm{argmax}}\limits_{n}(\mathbf{D}_{n}(i,j))}, \\ 
    \mathbf{D} = \sum\nolimits_k\mathbf{M}_k \odot \mathbf{D}_k,~~&~~
    \mathbf{A} = \sum\nolimits_k\mathbf{M}_k \odot \mathbf{A}_k, \\
    \mathbf{V} = \frac{1}{K}\sum\nolimits_k\mathbf{V}_k, ~~&~~
    \mathbf{L} = \frac{1}{K}\sum\nolimits_k\mathbf{L}_k. \\
\end{aligned}
\end{equation} 
These CG components are then fed into a differentiable renderer $\Lambda$~\cite{kato2018neural} to reconstruct an image:
\begin{align}\label{equ-render}
    \hat{\mathbf{I}} = \Lambda(\mathbf{D}, \mathbf{A}, \mathbf{V}, \mathbf{L}).
\end{align}
When training the network, we can minimize the distance between the input image $\mathbf{I}$ and the reconstructed image $\hat{\mathbf{I}}$ following the analysis-by-synthesis strategy~\cite{wu2020unsupervised}, so that the network parameters can be learned in an unsupervised manner. The probing and reconstruction process is carried out at three levels: low, middle, and high, with the receptive fields corresponding to $\frac{1}{4} \times$, $\frac{1}{2} \times$, and $1 \times$ of the image size. The bottom stage with $64 \times 64$ windows is not probed because the DNN has not yet well-comprehended images. More details are provided in the supplemental materials. 

\section{Experiments}
In the experiments, we examine three key steps in graphics probing to gain insights into the mechanisms behind representation construction: the generation of geometry $\mathbf{D}$ in Eqn.~\ref{equ-decompose}, the generation of camera view $\mathbf{V}$ in Eqn.~\ref{equ-decompose}, and the competition of graphics probes in Eqn.~\ref{equ-assemble}. Additionally, we delve into the conditions under which a 3D representation emerges.

\begin{figure*}
   \begin{center}
   \includegraphics[width=0.98\linewidth]{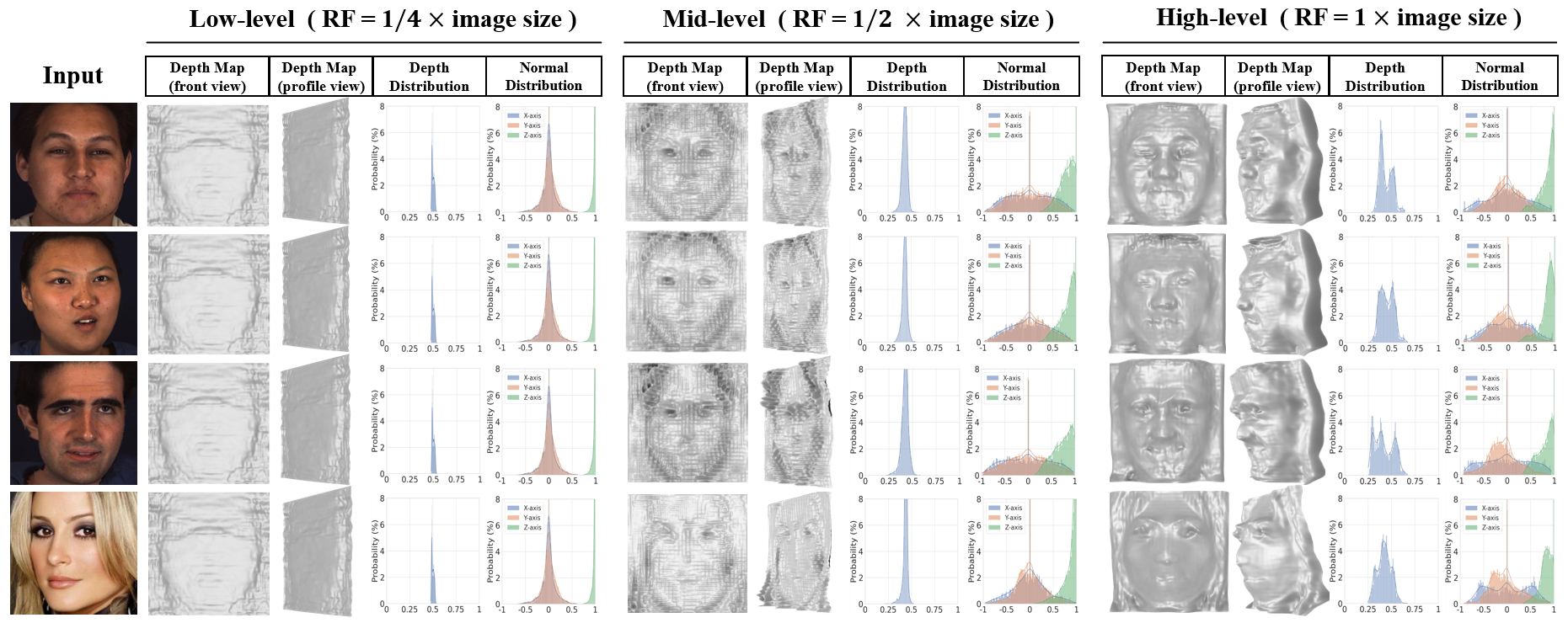}
   \caption{\textbf{Visualization of intermediate representations}. The geometry of representations at the low, middle, and high levels with receptive field (RF)  corresponding to $\frac{1}{4}\times$, $\frac{1}{2}\times$, and the full image size, respectively. At the low level, the geometry is flat, lacking any depth or normal variations. At the middle level, variations in normal begin to appear, yet the depth remains shallow, similar to a low-relief sculpture. At the high level, a fully 3D representation is constructed.} 
   \label{fig2-2D--2.5D--3D-vis}
   \end{center}
\end{figure*}

\subsection{Implementation Details}
We examine a 12-layer WinT, modified by the Swin-Tiny architecture~\cite{liu2021swin}. To analyze its representations, we insert probes at the 3rd (with $4 \times 4$ windows), 5th (with $2 \times 2$ windows), and 11th (with $1 \times 1$ windows) layers, corresponding to the low, middle and high levels, respectively. Our dataset encompasses both constrained and unconstrained scenarios.
The unconstrained scenario is derived from the CelebA dataset~\cite{liu2015deep}, which consists of $202,599$ images from $10,177$ distinct identities. The constrained scenario is created from the laser scans obtained from the BP4D dataset~\cite{zhang2014bp4d}. The dataset comprises $18$ male and $23$ female heads, rendered in $13$ yaw angles: $0^{\circ}$, $\pm15^{\circ}$, $\pm30^{\circ}$, $\pm45^{\circ}$, $\pm60^{\circ}$, $\pm75^{\circ}$, and $\pm90^{\circ}$, yielding a total of $19,376$ images. In the experiments, $90\%$ identities are used for training and the remaining $10\%$ are used for testing.

\subsection{Visualization of 2D--2.5D--3D Representations}
Figure~\ref{fig2-2D--2.5D--3D-vis} illustrates the probed depth map $\mathbf{D}$ in Eqn.~\ref{equ-decompose} at different levels, along with a statistical analysis of the depth and normal distributions for each sample. Key observations include:

\noindent \textbf{At the low level} with a receptive field of $\frac{1}{4}\times$ image size, we observe that the depth map resembles a 2D plane, which is further supported by the distribution of depth and normal direction.  The depth values are predominantly about $0.5$, and the normals are approximately $[0,0,1]$. This suggests that the low-level layers have 2D representations. Additionally, we have investigated lower layers with receptive fields of $\frac{1}{8}\times$ image size, which exhibit similar characteristics.

\noindent \textbf{At the middle level} with a receptive field of $\frac{1}{2}\times$ image size, the frontal views of depth maps clearly reveal the edges and contours of a human face. In contrast, the side views demonstrate that the reconstructed shapes remain nearly planar, similar to a low-relief sculpture. This implies that the network crafts surface normals to simulate shading effects. However, the depth variations are limited to a shallow range, meaning that true 3D structures are still not perceived. The distributions show little change in depth but a significant increase in the range of normal orientations compared to the lower levels. This observation aligns with Marr's theory of a 2.5D state in visual perception, which is characterized by the formation of surface orientations. 

\noindent \textbf{At the high level} with a receptive field covering the whole image, the side views of geometry indicate that a fully 3D shape is built. Moreover, the depth variations exhibit a significant increase compared to those at the low and middle levels. This observation implies that 3D shapes are probed in the top-layer representations, which is in accordance with Marr's theory that places 3D reconstruction at the end of visual perception.

The above observations are generally consistent across the images in the constrained and unconstrained scenarios. More illustrations are provided in the supplemental materials

\begin{figure*}
   \begin{center}
   \includegraphics[width=0.95\linewidth]{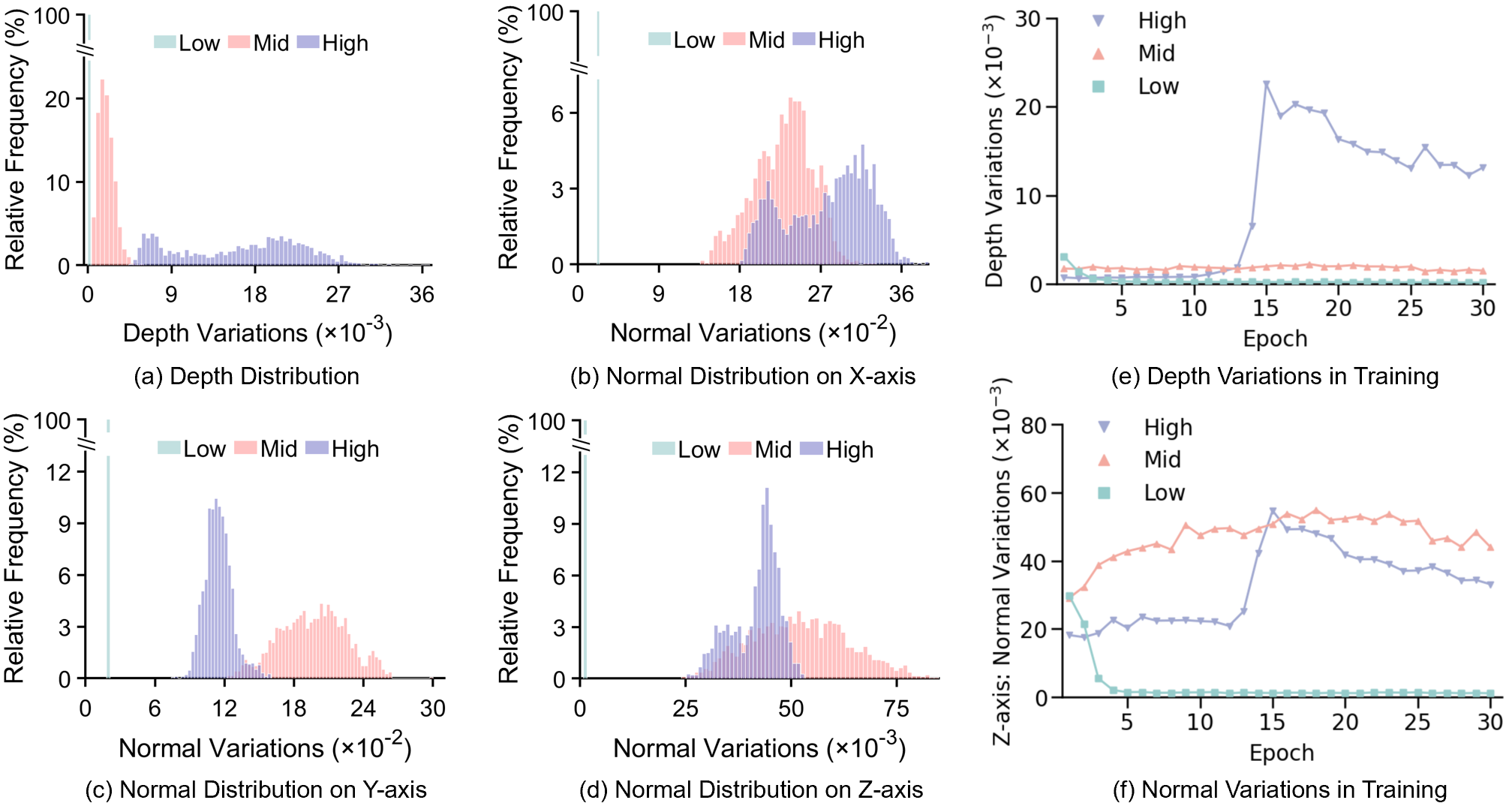}
   \caption{\textbf{Distribution of depth and normal variations}.  The distributions of variations for individual samples across the testing dataset for (a) depth, (b) x-axis of normal, (c) y-axis of normal, and (d)  z-axis of normal. The mean variations for (e) depth and (f) normal throughout the training process.} 
   \label{fig-distribution}
   \end{center}
\end{figure*}

\subsection{Distributions of Depth and Normal Variations}
We quantitatively evaluate the depth and normal variances across the dataset. We calculate the variations of the depth values and the $x, y, z$ of the normal vectors for each sample and then calculate the distribution of these variations within the test dataset. Figure~\ref{fig-distribution}(a) shows the depth variations. At the low level, the depth variations are zero, indicating a flat surface. When moving to the middle level, subtle depth variations begin to appear, indicating the presence of minor fluctuations. By examining the visualization in Figure~\ref{fig2-2D--2.5D--3D-vis}, we can conclude that these subtle fluctuations are responsible for producing the normal variations. At the high level, substantial depth variations are observed, which allows for the construction of a 3D shape. Besides, a clear distinction can be seen between 2.5D and 3D depth variations. Figure~\ref{fig-distribution}(b)-(d) depicts the normal variances, where it is evident that the variances are negligible at the low level and become significant at the middle and high levels. In summary, 2D representations exhibit low variances in both depth and normal, 2.5D representations show low depth variance but high normal variance, and 3D representations demonstrate high variance in both depth and normal.

To further investigate the learning process of these representations, we analyze the progression of depth and normal variances throughout the training process. Figure~\ref{fig-distribution}(e) illustrates that at the beginning of training, there are no depth variations, indicating an initial flat surface. Subsequently, a substantial increase in depth variations at the high level is observed, particularly around the $15$th epoch. Then, depth is continuously refined until the model converges. As for the normals, Figure \ref{fig-distribution}(f) reveals that there are variations from the start because of the noise introduced by random initialization. As the training progresses, the low-level normal variations diminish to zero, signifying a transition to a flatter geometry. The mid-level normal variations rise in a smooth manner. The high-level normal variations initially stay relatively low but then exhibit a sharp increase, corresponding to the pattern observed in the depth variations. From the analysis, we have three observations: 1) The sudden increase in depth and normal variations at the high level around the 15th epoch suggests that there is a critical point at which a 3D understanding of objects suddenly emerges. 2) The high-level normal variations remain low until the mid-level variations approach their peak. It appears that the high-level layers are awaiting insights from the middle level before the construction of a 3D representation. 3) The 2D representation is not a default state but rather an intentional result at the low level, as the network actively reduces the normal variations to a minimal extent, thus creating a flat geometry.

\subsection{From Viewer-centered to Object-centered}
\begin{figure}
   \begin{center}
   \includegraphics[width=0.98\linewidth]{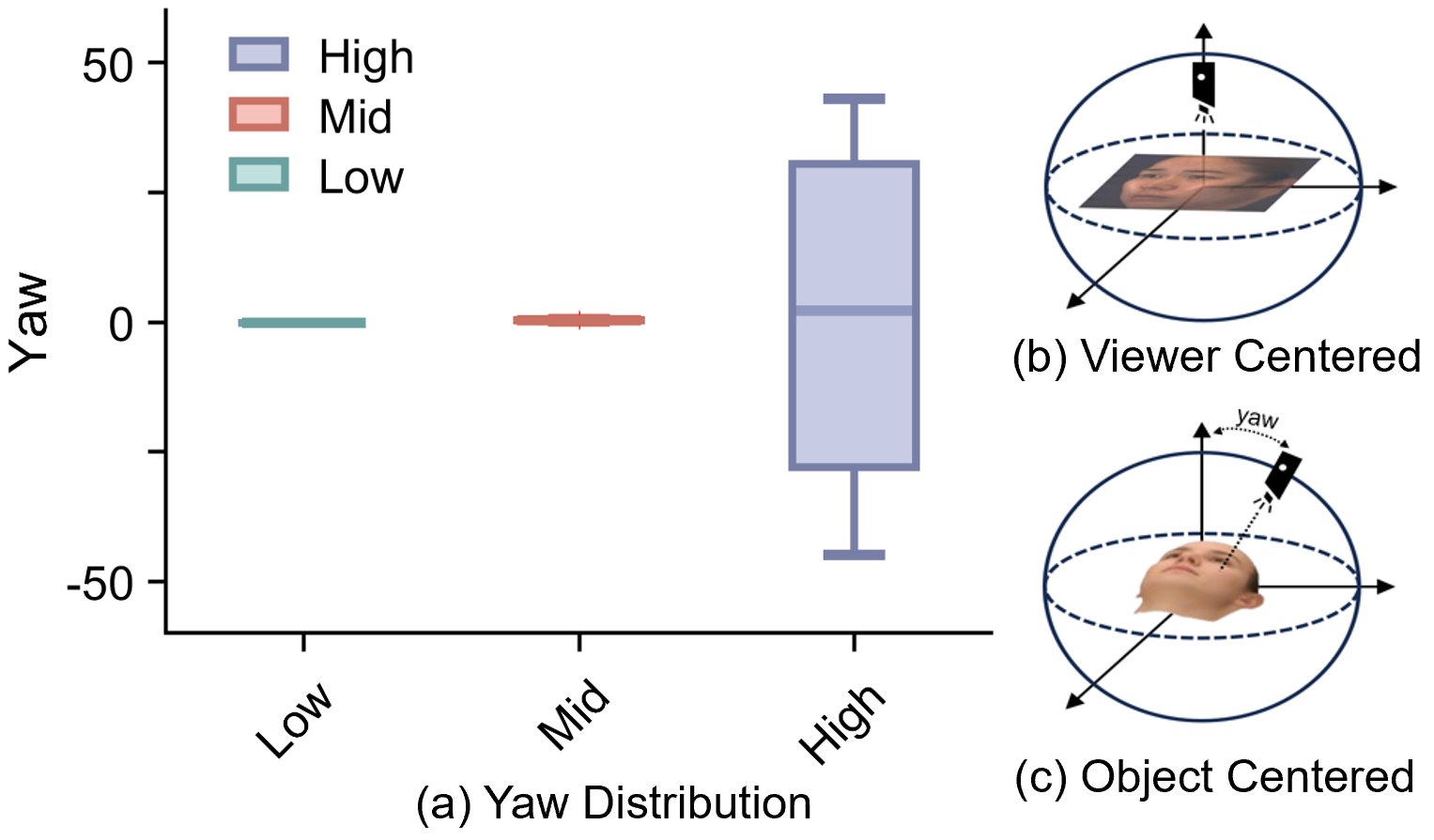}
   \caption{The distribution of yaw angles across different levels and an illustration of  viewer-centered and object-centered representations.}
   \label{fig-view}
   \end{center}
\end{figure}

There has been a longstanding debate~\cite{logothetis1995shape} on whether objects are fundamentally encoded with object-centered or viewer-centered representations.  In our experiments, by inserting probes into the intermediate layers, we obtain the 6DoF camera view $\mathbf{V}$ of the representation through Eqn.~\ref{equ-decompose}, thereby revealing the pitch, yaw, and roll of the view angle. Figure~\ref{fig-view}(a) illustrates the distribution of the perceived yaw angles, which exhibit the most significant variations in the dataset. It can be seen that the yaw angles at the low and middle levels are nearly zero. In these cases, the camera is consistently positioned above the image plane, with the viewing direction perpendicular to the image plane, as depicted in Figure~\ref{fig-view}(b). Since the world coordinates are anchored to the viewer's perspective, this configuration resembles a viewer-centered representation.
In contrast, the high level exhibits significant variations in yaw angles, primarily ranging between $[-50^\circ, 50^\circ]$. At this level, a frontal 3D face is placed at the center of a canonical space, and a disentangled camera view is introduced to render its profile, as illustrated in Figure~\ref{fig-view}(c). Since the world coordinates are anchored to the object's perspective, this configuration aligns with an object-centered representation.

\begin{figure}
   \begin{center}
   \includegraphics[width=0.98\linewidth]{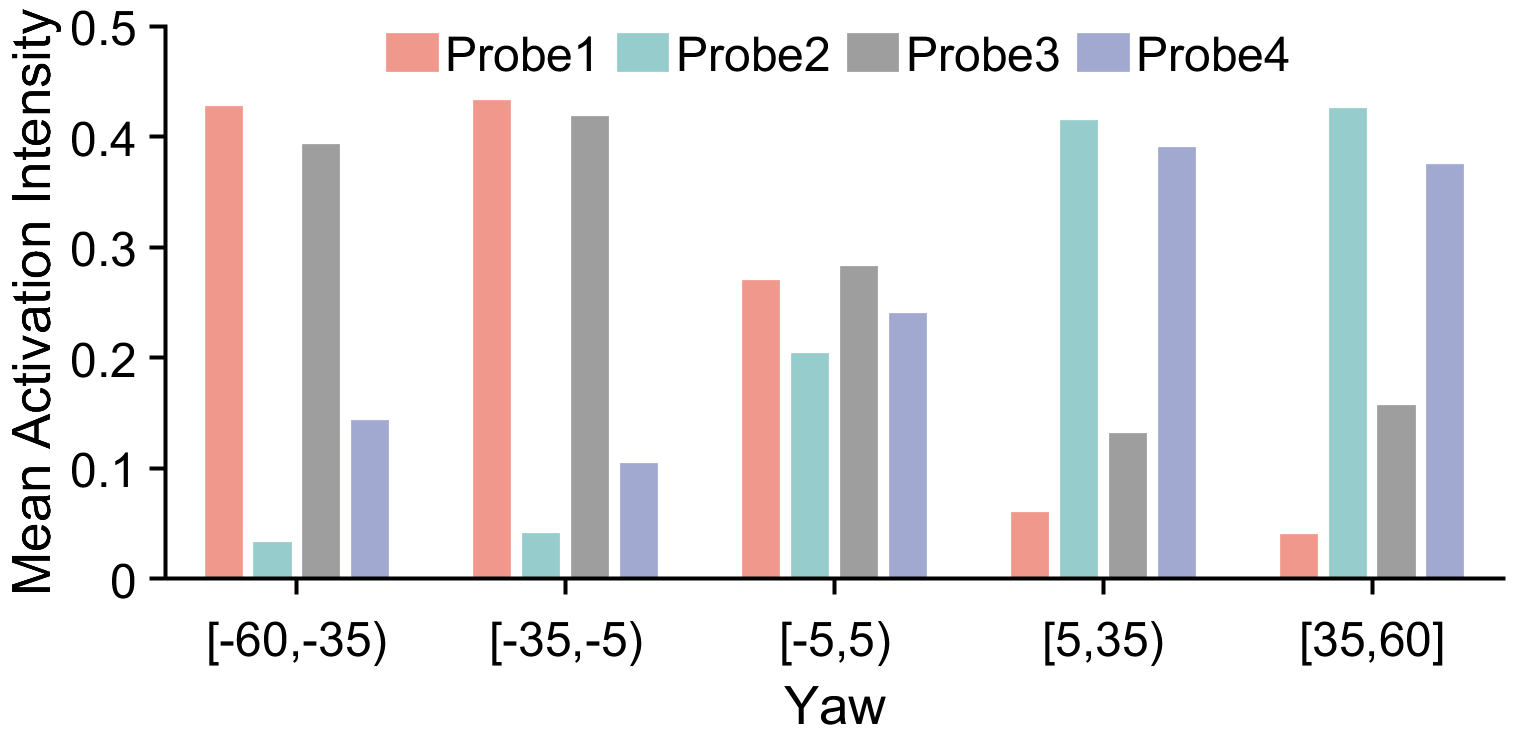}
   \caption{The view tuning at the middle level. The y-axis is the mean activation intensity, and the x-axis corresponds to the dataset subsets divided by the yaw angle.}
   \label{fig-tune-view}
   \end{center}
\end{figure}

\subsection{Tuning of Graphics Probe}
During the probing phase, each graphics probe $\mathbf{p}_{k}$ engages in a depth competition in Eqn.~\ref{equ-assemble}. If the depth map of $\mathbf{p}_{k}$ 
wins in this competition, its appearance is either fully or partially revealed in the final reconstructed image. In such cases, we consider that the input image activates $\mathbf{p}_{k}$, or, in the context of neuroscience, $\mathbf{p}_{k}$ is tuned to the input image.

\begin{figure*}
   \begin{center}
   \includegraphics[width=0.98\linewidth]{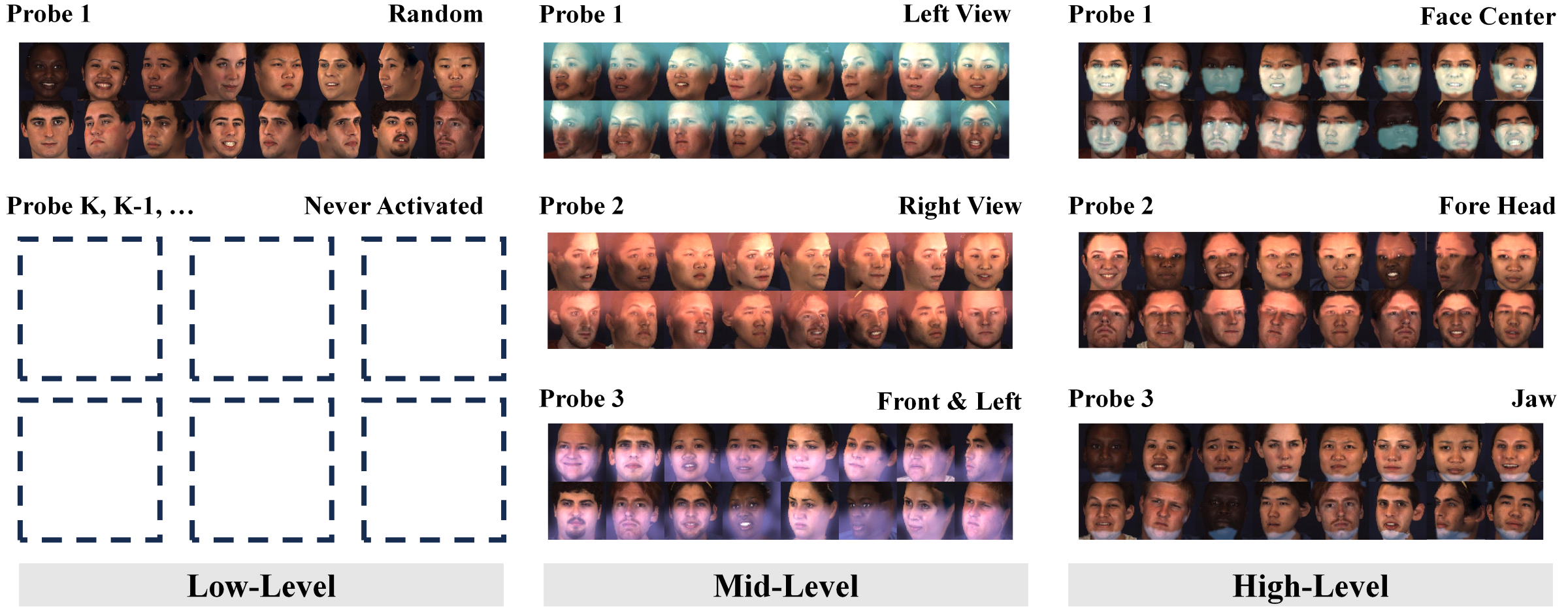}
   \caption{\textbf{Tuning of the graphics probes.} The samples that activate specific graphics probes, with the activated regions highlighted. At the low level, the activated samples reveal no clear semantic meanings. At the middle level, the probes exhibit a view-tuning behavior, with certain probes primarily responding to the left-view and right-view samples. At the high level, the probes display a part-tuning behavior, focusing on specific facial components such as the face center, forehead, and jaw.}
   \label{fig-tune}
   \end{center}
\end{figure*}

For each graphics probe, we collect the images that activate it and display these images in Figure~\ref{fig-tune}. It can be observed that: 1) At the low level, where the representations are 2D, the tuning mechanism does not exhibit clear patterns. Besides, we find only a minority of probes are frequently activated, while the majority remain never activated.  2) At the middle level, where the representations are 2.5D, the probes show a tendency towards view-tuning behavior. For instance, most images that activate the first probe are left-view faces, while those that activate the second probe are predominantly right-view faces. 3) At the high level, probes are tuned to all views but concentrate on different regions. These regions consistently correspond to specific facial components, such as the face center, forehead, and jaw. This indicates that the tuning has become view-invariant and has shifted to part-tuning. Besides, this part-decomposition mechanism suggests that the network has developed a hierarchical understanding of faces.

To quantitatively analyze the view-tuning behavior at the middle level, we define the activation intensity of a probe to an image as the ratio of the probe's activation area to the total area of image. Additionally, the test dataset is divided into different yaw intervals: $[-60^\circ,-35^\circ]$, $[-35^\circ,-5^\circ]$, $[-5^\circ,5^\circ]$, $[5^\circ,35^\circ]$, and $[35^\circ,60^\circ]$. In Figure~\ref{fig-tune-view}, we show the activation intensities for four mid-level probes across these intervals.  It is observed that for yaw angles below $-5^\circ$, the network primarily activates Probe1 and Probe3, which are more sensitive to left-view faces. For yaw angles above $5^\circ$, it tends to activate Probe2 and Probe4, which are more sensitive to right-view faces. At yaw angles near $0^\circ$, the activation is evenly distributed across all probes. This observation further validates that the mid-level layers are sensitive to changes in viewpoint.

\begin{figure}
   \begin{center}
   \includegraphics[width=0.80\linewidth]{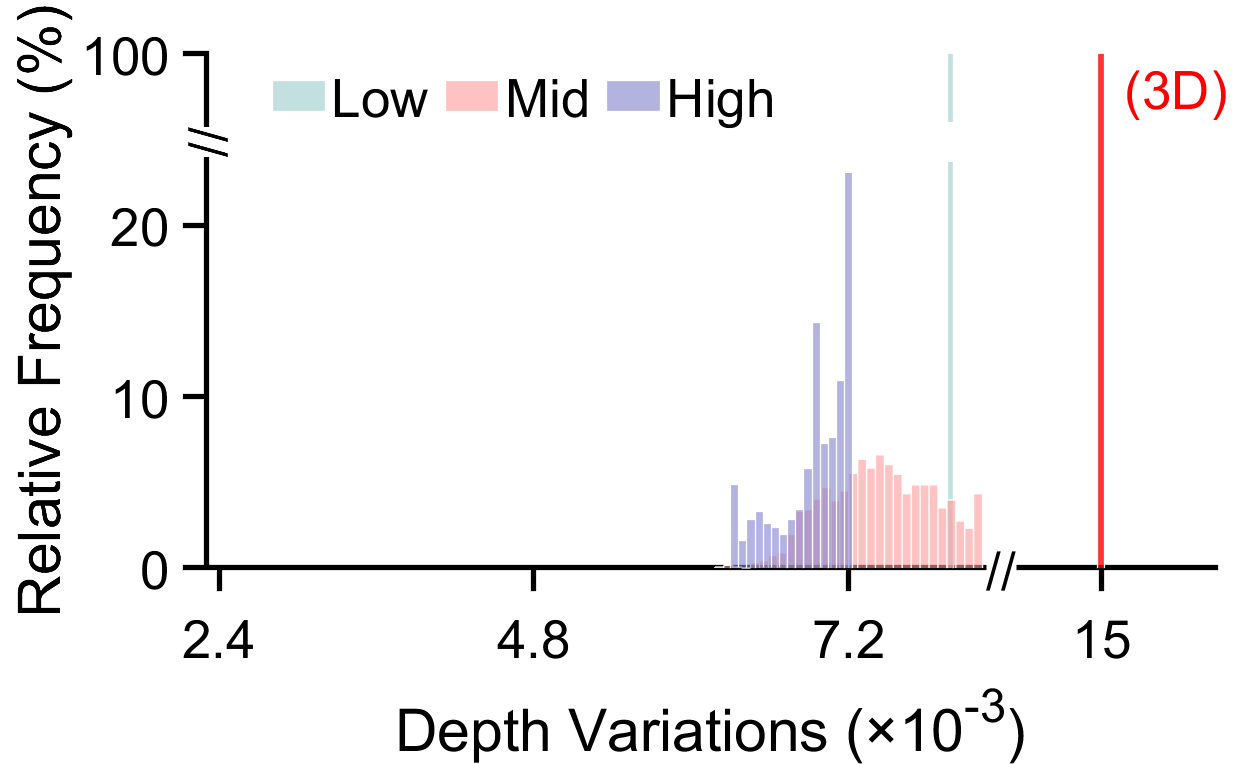}
   \caption{Depth variations distribution when trained on the dataset without view variations. The threshold necessary for 3D geometry is set at $15 \times 10^{-3}$.}
   \label{fig-3d-emergence}
   \end{center}
\end{figure}

\subsection{The Emergence of 3D Representation}
The conditions under which a 3D representation is constructed is an important question. If an object is inherently 2D and lacks any viewpoint variations, will a 3D representation still be formed? To simulate this scenario, we train a network on a dataset consisting of only frontal faces, with the results presented in Figure~\ref{fig-3d-emergence}. It is evident that when trained exclusively with images from a single viewpoint, the learned representations demonstrate very shallow depth spans. Even at the high level, the depth variations do not reach the threshold necessary for 3D geometry, indicating a failure to learn 3D shapes. This suggests that the network lacks the incentive to construct 3D representations when there are no appearance variations caused by viewpoint changes. If the object is inherently 2D, a 2D representation is sufficient. A comprehensive 3D surround view of the object is a prerequisite for the emergence of 3D representations. Additional illustrations are shown in the supplemental materials. 

\subsection{Analysis on More Architectures}
In addition to the WinT, we examine more network architectures to determine whether the construction of 2D, 2.5D, and 3D representations is a common feature across different architectures. The additional architectures include VGG16~\cite{simonyan2014very}, Resnet18~\cite{he2016deep}, Swin Transformer (SwinT)~\cite{liu2021swin}, and Vision Transformer (ViT)~\cite{dosovitskiy2020image}. For SwinT and ViT, we insert probe tokens into the first layer of each block and convert them into graphical probes in the last layer of the block, following the approach used in WinT. For VGG16 and Resnet18, we transform the last-layer feature of each block into graphical probes to visualize the representations. As shown in Table~\ref{tab-architecture}, across all architectures, the low-level representations exhibit small depth and normal variances, the mid-level representations display small depth with large normal variances, and the high-level representations have both large depth and normal variances, corresponding to the 2D, 2.5D, and 3D representations, respectively. The visualization are provided in the supplemental materials, which also adhere to the 2D--2.5D--3D mechanism. This evaluation confirms the generalizability of Marr's theory across various network architectures.

\begin{table}
    \centering
    \caption{Evaluation on more architectures. The mean variations of depth and z-axis of normal$(\times10^{-3})$ are demonstrated.}
    \label{tab-architecture}
      \resizebox{0.50\textwidth}{!}{
      \begin{tabular}{ccccccc}
      \toprule[1pt]
      \multirow{2}{*}[-0.5ex]{Network}  & \multicolumn{2}{c}{Low-Level} & \multicolumn{2}{c}{Mid-Level} & \multicolumn{2}{c}{High-Level}\\
      \cmidrule(lr){2-3} \cmidrule(lr){4-5}  \cmidrule(lr){6-7}    & Depth & Normal  & Depth & Normal & Depth & Normal  \\
      \midrule[1pt]
      VGG16~\cite{simonyan2014very}  &  0.210 & 2.71  & 2.59  &  29.8 &  15.4 & 59.6 \\
      Resnet18~\cite{he2016deep}   &  0.262 & 2.93  & 1.12  &  26.4 &  18.4 & 63.7 \\      
      ViT~\cite{dosovitskiy2020image}   &  0.178 & 1.67  & 0.759 & 35.3 & 18.7  & 43.3 \\
      SwinT~\cite{liu2021swin}   &  0.188 & 1.92  & 1.93 & 31.0 &  18.1 & 64.4 \\
      WinT   &  0.209 & 1.16  & 1.99  &  52.4 &  16.3 & 41.8   \\
      \bottomrule[0.75pt]
      \end{tabular}
      }
      
\end{table}%

\section{Conclusion}
In this paper, we introduce graphics probe, a new approach that effectively converts a network's intermediate feature into visualizable computer graphics (CG) elements, including depth, albedo, camera view, and lighting. Our analysis of the probed depth indicates that DNNs initially form 2D representations, then evolve to 2.5D representations that capture surface normals with limited depth, and finally build 3D shapes. This sequential progression from 2D to 2.5D to 3D is consistent with David Marr's seminal theory of vision. Furthermore, we observe phenomenons that indicate features at lower levels are viewer-centered and tuned to specific viewpoints, whereas high-level features are object-centered and tuned to facial components, which provides new insights into the role of viewpoint disentanglement in perception. Finally, we find that observing objects from diverse 3D viewpoints is a prerequisite for constructing a 3D representation.

\small
\bibliographystyle{IEEEtran}
\bibliography{sn-bibliography}

\begin{thebibliography}{10}
\providecommand{\url}[1]{#1}
\csname url@samestyle\endcsname
\providecommand{\newblock}{\relax}
\providecommand{\bibinfo}[2]{#2}
\providecommand{\BIBentrySTDinterwordspacing}{\spaceskip=0pt\relax}
\providecommand{\BIBentryALTinterwordstretchfactor}{4}
\providecommand{\BIBentryALTinterwordspacing}{\spaceskip=\fontdimen2\font plus
\BIBentryALTinterwordstretchfactor\fontdimen3\font minus \fontdimen4\font\relax}
\providecommand{\BIBforeignlanguage}[2]{{%
\expandafter\ifx\csname l@#1\endcsname\relax
\typeout{** WARNING: IEEEtran.bst: No hyphenation pattern has been}%
\typeout{** loaded for the language `#1'. Using the pattern for}%
\typeout{** the default language instead.}%
\else
\language=\csname l@#1\endcsname
\fi
#2}}
\providecommand{\BIBdecl}{\relax}
\BIBdecl

\bibitem{marr1978representation}
D.~Marr and H.~K. Nishihara, ``Representation and recognition of the spatial organization of three-dimensional shapes,'' \emph{Proceedings of the Royal Society of London. Series B. Biological Sciences}, vol. 200, no. 1140, pp. 269--294, 1978.

\bibitem{logothetis1995shape}
N.~K. Logothetis, J.~Pauls, and T.~Poggio, ``Shape representation in the inferior temporal cortex of monkeys,'' \emph{Current biology}, vol.~5, no.~5, pp. 552--563, 1995.

\bibitem{poggio2004generalization}
T.~Poggio and E.~Bizzi, ``Generalization in vision and motor control,'' \emph{Nature}, vol. 431, no. 7010, pp. 768--774, 2004.

\bibitem{diwadkar1997viewpoint}
V.~A. Diwadkar and T.~P. McNamara, ``Viewpoint dependence in scene recognition,'' \emph{Psychological science}, vol.~8, no.~4, pp. 302--307, 1997.

\bibitem{yamins2016using}
D.~L. Yamins and J.~J. DiCarlo, ``Using goal-driven deep learning models to understand sensory cortex,'' \emph{Nature neuroscience}, vol.~19, no.~3, pp. 356--365, 2016.

\bibitem{bansal2016marr}
A.~Bansal, B.~Russell, and A.~Gupta, ``Marr revisited: 2d-3d alignment via surface normal prediction,'' in \emph{Proceedings of the IEEE conference on computer vision and pattern recognition}, 2016, pp. 5965--5974.

\bibitem{yildirim2020efficient}
I.~Yildirim, M.~Belledonne, W.~Freiwald, and J.~Tenenbaum, ``Efficient inverse graphics in biological face processing,'' \emph{Science advances}, vol.~6, no.~10, p. eaax5979, 2020.

\bibitem{tacchetti2018invariant}
A.~Tacchetti, L.~Isik, and T.~A. Poggio, ``Invariant recognition shapes neural representations of visual input,'' \emph{Annual review of vision science}, vol.~4, pp. 403--422, 2018.

\bibitem{kazhdan2003rotation}
M.~Kazhdan, T.~Funkhouser, and S.~Rusinkiewicz, ``Rotation invariant spherical harmonic representation of 3 d shape descriptors,'' in \emph{Symposium on geometry processing}, vol.~6, 2003, pp. 156--164.

\bibitem{liebelt2008independent}
J.~Liebelt, C.~Schmid, and K.~Schertler, ``independent object class detection using 3d feature maps,'' in \emph{2008 IEEE Conference on Computer Vision and Pattern Recognition}.\hskip 1em plus 0.5em minus 0.4em\relax IEEE, 2008, pp. 1--8.

\bibitem{liu2020recognizing}
S.~Liu, V.~Nguyen, I.~Rehg, and Z.~Tu, ``Recognizing objects from any view with object and viewer-centered representations,'' in \emph{Proceedings of the IEEE/CVF Conference on Computer Vision and Pattern Recognition}, 2020, pp. 11\,784--11\,793.

\bibitem{calder2011oxford}
A.~J. Calder, \emph{Oxford handbook of face perception}.\hskip 1em plus 0.5em minus 0.4em\relax Oxford University Press, USA, 2011.

\bibitem{logothetis1994view}
N.~Logothetis, J.~Pauls, H.~B{\"u}lthoff, and T.~Poggio, ``View-dependent object recognition by monkeys,'' \emph{Current biology}, vol.~4, no.~5, pp. 401--414, 1994.

\bibitem{masi2018deep}
I.~Masi, Y.~Wu, T.~Hassner, and P.~Natarajan, ``Deep face recognition: A survey,'' in \emph{2018 31st SIBGRAPI conference on graphics, patterns and images (SIBGRAPI)}.\hskip 1em plus 0.5em minus 0.4em\relax IEEE, 2018, pp. 471--478.

\bibitem{wu2019facial}
Y.~Wu and Q.~Ji, ``Facial landmark detection: A literature survey,'' \emph{International Journal of Computer Vision}, vol. 127, pp. 115--142, 2019.

\bibitem{li2020deep}
S.~Li and W.~Deng, ``Deep facial expression recognition: A survey,'' \emph{IEEE transactions on affective computing}, vol.~13, no.~3, pp. 1195--1215, 2020.

\bibitem{hill2019deep}
M.~Q. Hill, C.~J. Parde, C.~D. Castillo, Y.~I. Colon, R.~Ranjan, J.-C. Chen, V.~Blanz, and A.~J. Ooole, ``Deep convolutional neural networks in the face of caricature,'' \emph{Nature Machine Intelligence}, vol.~1, no.~11, pp. 522--529, 2019.

\bibitem{o2018face}
A.~J. Ooole, C.~D. Castillo, C.~J. Parde, M.~Q. Hill, and R.~Chellappa, ``Face space representations in deep convolutional neural networks,'' \emph{Trends in cognitive sciences}, vol.~22, no.~9, pp. 794--809, 2018.

\bibitem{parde2017face}
C.~J. Parde, C.~Castillo, M.~Q. Hill, Y.~I. Colon, S.~Sankaranarayanan, J.-C. Chen, and A.~J. Ooole, ``Face and image representation in deep cnn features,'' in \emph{2017 12th IEEE International Conference on Automatic Face \& Gesture Recognition (FG 2017)}.\hskip 1em plus 0.5em minus 0.4em\relax IEEE, 2017, pp. 673--680.

\bibitem{vaswani2017attention}
A.~Vaswani, N.~Shazeer, N.~Parmar, J.~Uszkoreit, L.~Jones, A.~N. Gomez, {\L}.~Kaiser, and I.~Polosukhin, ``Attention is all you need,'' \emph{Advances in neural information processing systems}, vol.~30, 2017.

\bibitem{kim2018interpretability}
B.~Kim, M.~Wattenberg, J.~Gilmer, C.~Cai, J.~Wexler, F.~Viegas \emph{et~al.}, ``Interpretability beyond feature attribution: Quantitative testing with concept activation vectors (tcav),'' in \emph{International conference on machine learning}.\hskip 1em plus 0.5em minus 0.4em\relax PMLR, 2018, pp. 2668--2677.

\bibitem{alain2016understanding}
G.~Alain and Y.~Bengio, ``Understanding intermediate layers using linear classifier probes,'' in \emph{International Conference on Learning Representations}, 2017, pp. 1542--1553.

\bibitem{zhong2016face}
Y.~Zhong, J.~Sullivan, and H.~Li, ``Face attribute prediction using off-the-shelf cnn features,'' in \emph{2016 International Conference on Biometrics (ICB)}.\hskip 1em plus 0.5em minus 0.4em\relax IEEE, 2016, pp. 1--7.

\bibitem{terhorst2021soft}
P.~Terh{\"o}rst, D.~F{\"a}hrmann, N.~Damer, F.~Kirchbuchner, and A.~Kuijper, ``On soft-biometric information stored in biometric face embeddings,'' \emph{IEEE Transactions on Biometrics, Behavior, and Identity Science}, vol.~3, no.~4, pp. 519--534, 2021.

\bibitem{dhar2020attributes}
P.~Dhar, A.~Bansal, C.~D. Castillo, J.~Gleason, P.~J. Phillips, and R.~Chellappa, ``How are attributes expressed in face dcnns?'' in \emph{2020 15th IEEE International Conference on Automatic Face and Gesture Recognition (FG 2020)}.\hskip 1em plus 0.5em minus 0.4em\relax IEEE, 2020, pp. 85--92.

\bibitem{yu2023graphics}
C.~Yu, X.~Zhu, X.~Zhang, Z.~Zhang, and Z.~Lei, ``Graphics capsule: learning hierarchical 3d face representations from 2d images,'' in \emph{Proceedings of the IEEE/CVF Conference on Computer Vision and Pattern Recognition}, 2023, pp. 20\,981--20\,990.

\bibitem{fong2018net2vec}
R.~Fong and A.~Vedaldi, ``Net2vec: Quantifying and explaining how concepts are encoded by filters in deep neural networks,'' in \emph{Proceedings of the IEEE conference on computer vision and pattern recognition}, 2018, pp. 8730--8738.

\bibitem{grossman2019convergent}
S.~Grossman, G.~Gaziv, E.~Yeagle, M.~Harel, P.~Mégevand, D.~M. Groppe, S.~Khuvis, J.~Herrero, M.~Irani, A.~Mehta, and R.~Malach, ``Convergent evolution of face spaces across human face-selective neuronal groups and deep convolutional networks,'' \emph{Nature Communications}, vol.~10, 2019.

\bibitem{wu2017marrnet}
J.~Wu, Y.~Wang, T.~Xue, X.~Sun, B.~Freeman, and J.~Tenenbaum, ``Marrnet: 3d shape reconstruction via 2.5 d sketches,'' \emph{Advances in neural information processing systems}, vol.~30, 2017.

\bibitem{sun2018pix3d}
X.~Sun, J.~Wu, X.~Zhang, Z.~Zhang, C.~Zhang, T.~Xue, J.~B. Tenenbaum, and W.~T. Freeman, ``Pix3d: Dataset and methods for single-image 3d shape modeling,'' in \emph{Proceedings of the IEEE conference on computer vision and pattern recognition}, 2018, pp. 2974--2983.

\bibitem{lun20173d}
Z.~Lun, M.~Gadelha, E.~Kalogerakis, S.~Maji, and R.~Wang, ``3d shape reconstruction from sketches via multi-view convolutional networks,'' in \emph{2017 International Conference on 3D Vision (3DV)}.\hskip 1em plus 0.5em minus 0.4em\relax IEEE, 2017, pp. 67--77.

\bibitem{wu2018learning}
J.~Wu, C.~Zhang, X.~Zhang, Z.~Zhang, W.~T. Freeman, and J.~B. Tenenbaum, ``Learning shape priors for single-view 3d completion and reconstruction,'' in \emph{Proceedings of the European conference on computer vision (ECCV)}, 2018, pp. 646--662.

\bibitem{zhang2018learning}
X.~Zhang, Z.~Zhang, C.~Zhang, J.~Tenenbaum, B.~Freeman, and J.~Wu, ``Learning to reconstruct shapes from unseen classes,'' \emph{Advances in neural information processing systems}, vol.~31, 2018.

\bibitem{yu2022degenerate}
T.~Yu and P.~Li, ``Degenerate swin to win: Plain window-based transformer without sophisticated operations,'' \emph{arXiv preprint arXiv:2211.14255}, 2022.

\bibitem{liu2021swin}
Z.~Liu, Y.~Lin, Y.~Cao, H.~Hu, Y.~Wei, Z.~Zhang, S.~Lin, and B.~Guo, ``Swin transformer: Hierarchical vision transformer using shifted windows,'' in \emph{Proceedings of the IEEE/CVF international conference on computer vision}, 2021, pp. 10\,012--10\,022.

\bibitem{dosovitskiy2020image}
A.~Dosovitskiy, L.~Beyer, A.~Kolesnikov, D.~Weissenborn, X.~Zhai, T.~Unterthiner, M.~Dehghani, M.~Minderer, G.~Heigold, S.~Gelly, J.~Uszkoreit, and N.~Houlsby, ``An image is worth 16x16 words: Transformers for image recognition at scale,'' in \emph{International Conference on Learning Representations}, 2021, pp. 556--577.

\bibitem{kato2018neural}
H.~Kato, Y.~Ushiku, and T.~Harada, ``Neural 3d mesh renderer,'' in \emph{Proceedings of the IEEE conference on computer vision and pattern recognition}, 2018, pp. 3907--3916.

\bibitem{wu2020unsupervised}
S.~Wu, C.~Rupprecht, and A.~Vedaldi, ``Unsupervised learning of probably symmetric deformable 3d objects from images in the wild,'' in \emph{Proceedings of the IEEE/CVF Conference on Computer Vision and Pattern Recognition}, 2020, pp. 1--10.

\bibitem{liu2015deep}
Z.~Liu, P.~Luo, X.~Wang, and X.~Tang, ``Deep learning face attributes in the wild,'' in \emph{Proceedings of the IEEE international conference on computer vision}, 2015, pp. 3730--3738.

\bibitem{zhang2014bp4d}
X.~Zhang, L.~Yin, J.~F. Cohn, S.~Canavan, M.~Reale, A.~Horowitz, P.~Liu, and J.~M. Girard, ``Bp4d-spontaneous: a high-resolution spontaneous 3d dynamic facial expression database,'' \emph{Image and Vision Computing}, vol.~32, no.~10, pp. 692--706, 2014.

\bibitem{simonyan2014very}
K.~Simonyan and A.~Zisserman, ``Very deep convolutional networks for large-scale image recognition,'' in \emph{International Conference on Learning Representations}, 2015, pp. 1--14.

\bibitem{he2016deep}
K.~He, X.~Zhang, S.~Ren, and J.~Sun, ``Deep residual learning for image recognition,'' in \emph{Proceedings of the IEEE conference on computer vision and pattern recognition}, 2016, pp. 770--778.

\end{thebibliography}


\begin{thebibliography}{1}
\providecommand{\url}[1]{#1}
\csname url@samestyle\endcsname
\providecommand{\newblock}{\relax}
\providecommand{\bibinfo}[2]{#2}
\providecommand{\BIBentrySTDinterwordspacing}{\spaceskip=0pt\relax}
\providecommand{\BIBentryALTinterwordstretchfactor}{4}
\providecommand{\BIBentryALTinterwordspacing}{\spaceskip=\fontdimen2\font plus
\BIBentryALTinterwordstretchfactor\fontdimen3\font minus \fontdimen4\font\relax}
\providecommand{\BIBforeignlanguage}[2]{{%
\expandafter\ifx\csname l@#1\endcsname\relax
\typeout{** WARNING: IEEEtran.bst: No hyphenation pattern has been}%
\typeout{** loaded for the language `#1'. Using the pattern for}%
\typeout{** the default language instead.}%
\else
\language=\csname l@#1\endcsname
\fi
#2}}
\providecommand{\BIBdecl}{\relax}
\BIBdecl

\bibitem{liu2021swin}
Z.~Liu, Y.~Lin, Y.~Cao, H.~Hu, Y.~Wei, Z.~Zhang, S.~Lin, and B.~Guo, ``Swin transformer: Hierarchical vision transformer using shifted windows,'' in \emph{Proceedings of the IEEE/CVF international conference on computer vision}, 2021, pp. 10\,012--10\,022.

\bibitem{wu2020unsupervised}
S.~Wu, C.~Rupprecht, and A.~Vedaldi, ``Unsupervised learning of probably symmetric deformable 3d objects from images in the wild,'' in \emph{Proceedings of the IEEE/CVF Conference on Computer Vision and Pattern Recognition}, 2020, pp. 1--10.

\bibitem{simonyan2014very}
K.~Simonyan and A.~Zisserman, ``Very deep convolutional networks for large-scale image recognition,'' in \emph{International Conference on Learning Representations}, 2015, pp. 1--14.

\bibitem{he2016deep}
K.~He, X.~Zhang, S.~Ren, and J.~Sun, ``Deep residual learning for image recognition,'' in \emph{Proceedings of the IEEE conference on computer vision and pattern recognition}, 2016, pp. 770--778.

\bibitem{yu2022degenerate}
T.~Yu and P.~Li, ``Degenerate swin to win: Plain window-based transformer without sophisticated operations,'' \emph{arXiv preprint arXiv:2211.14255}, 2022.

\bibitem{dosovitskiy2020image}
A.~Dosovitskiy, L.~Beyer, A.~Kolesnikov, D.~Weissenborn, X.~Zhai, T.~Unterthiner, M.~Dehghani, M.~Minderer, G.~Heigold, S.~Gelly, J.~Uszkoreit, and N.~Houlsby, ``An image is worth 16x16 words: Transformers for image recognition at scale,'' in \emph{International Conference on Learning Representations}, 2021, pp. 556--577.

\end{thebibliography}

\end{document}


\maketitle
\hyphenpenalty=5000
\tolerance=1000

\section{Experimental Details}
\subsection{Dataset}
\begin{figure*}
   \begin{center}
   \includegraphics[width=0.98\linewidth]{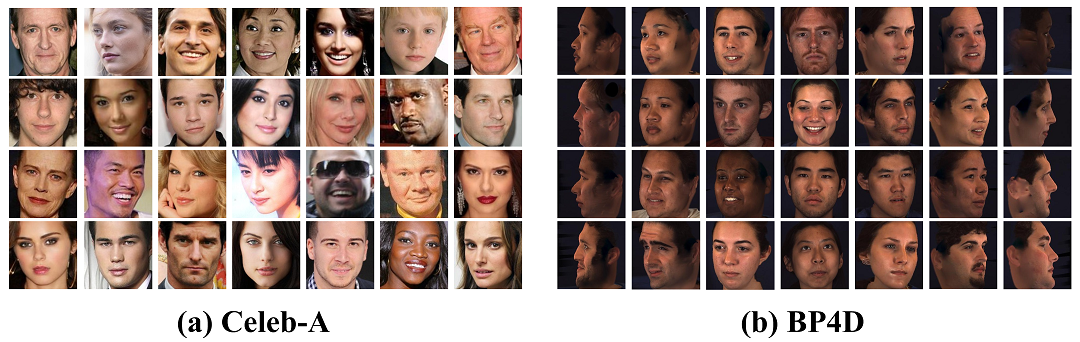}
   \caption{The datasets used in the experiments. (a) The Celeb-A dataset, which is collected in the unconstrained scenarios. (b) The BP4D dataset,  which is created by rendering 3D scans across extensive yaw angles.}
   \label{fig-dataset}
   \end{center}
\end{figure*}

Our experiments are performed on a combination of two datasets, each collected under different conditions: constrained and unconstrained scenarios. The unconstrained dataset is the CelebA, comprising $202,599$ images from $10,177$ unique individuals, as depicted in Figure~\ref{fig-dataset}(a). This dataset offers a rich variety of views and is captured in complex environments. However, the variation in views is limited, with the majority of the faces being frontal. Considering the strong correlation between our research and view variations, we also introduce a dataset that provides a more controlled environment with substantial pose variations, particularly in yaw angle, which is of significant interest in neuroscience. To achieve this, we have incorporated a dataset derived from laser scans of the BP4D dataset. This dataset features $18$ male and $23$ female heads, rendered from 13 different viewpoints (yaw: $0^{\circ}$, $\pm15^{\circ}$, $\pm30^{\circ}$, $\pm45^{\circ}$, $\pm60^{\circ}$, $\pm75^{\circ}$, $\pm90^{\circ}$), resulting in a total of $19,376$ images after cleaning. These images are displayed in Figure~\ref{fig-dataset}(b). In the experiments, $90\%$ identities from both datasets are used for training, and the rest $10\%$ are used for testing.

\subsection{Architecture}
Most of the Window Transformer (WinT) architecture follows the Swin-tiny architecture in the Swin Transformer (SwinT)~\cite{liu2021swin}. The only difference between our WinT and SwinT is that we remove the shifted windowing configuration to achieve a controlled receptive field for each token. Specifically, the $224 \times 224$ input face image is first split into non-overlapping $4 \times 4$ patches by a patch splitting module, and each patch is treated as a token, resulting in a feature map of size $56 \times 56$. Subsequently, this token grid is partitioned into windows of $7 \times 7$ tokens, resulting in a total of $8 \times 8$ windows. These windows serve as the input to the main architecture. The architecture hyper-parameters are listed in Table~\ref{tab-architecture}.

\begin{table*}
    \centering
    \caption{The hyper-parameters and characteristics of each stage. $H$ represents the height of the image, which is equal to the width. ``GP Num.'' is the number of graphics probes.}
      \resizebox{0.75\textwidth}{!}{
      \begin{tabular}{ccccccc}
      \toprule[1pt]
      \multirow{2}{*}[-0.5ex]{Stages}  & \multicolumn{4}{c}{Hyper-Parameters} & \multicolumn{2}{c}{Characteristics} \\
      \cmidrule(lr){2-5} \cmidrule(lr){6-7}   & Blocks & Window Num. & Rceptive Field  & GP Num. & Dimension & Tuning  \\
      \midrule[1pt]
      Bottom & $2$ & $64$ ($8 \times 8$) & $1/8 \times H$ & $64$ & 2D & Random \\
      Low & $2$ & $16$ ($4 \times 4$) & $1/4 \times H$ & $16$ & 2D & Random \\
      Mid & $6$ & $4$ ($2 \times 2$) & $1/2 \times H$ & $4$ & 2.5D & View \\
      High & $2$ & $1$ &  $H$ & $6$ & 3D & Components \\
      \bottomrule[0.75pt]
      \end{tabular}
      }
      \label{tab-architecture}
\end{table*}%

The main architecture is built by four stages, each comprising a specific number of WinT blocks: $[2,2,6,2]$. Within each WinT block, there are two consecutive multi-head self-attention modules that operate on the tokens within each window. After each stage, the number of tokens is reduced by a patch merging layer. The patch merging layer fuses the features of each group of $2 \times 2$ neighboring patches, reducing the number of tokens by a factor of $2 \times 2 = 4$ (a $2 \times 2$ downsampling of resolution). Therefore, the number of windows of the four stages are $8 \times 8$, $4 \times 4$, $2 \times 2$, and $1 \times 1$, and the corresponding receptive fields being $1/8 \times$, $1/4 \times$, $1/2 \times$, and $1 \times$ image size. The receptive field is equivalent to the window size, as each token can only access information within its window.

In each stage of our architecture, we maintain an additional token for every window, with the same dimension as the regular tokens. This additional token is inserted at the beginning of each stage and participates in the subsequent self-attention operations. At the end of each stage, these additional tokens serve as probe tokens, and are transformed into depth maps, albedo maps, lighting, and view by the graphics probes. It's important to note that the template activation, as described by Eqn.2 in the main text, varies slightly at the high level. At the low and middle levels, each probe token generates a single graphics probe, meaning the number of graphics probes is equal to the number of probe tokens. However, at the high level, there is only one window, and consequently, only one probe token is present. In this case, we still maintain $6$ templates, and replicate the single probe token $6$ times to match these templates, creating six distinct graphics probes as the final output. This operation helps us find the component tuning mechanism at the top layer of neural networks.

\subsection{Training}
When training the network with unlabelled images, we adopt the negative log-likelihood loss~\cite{wu2020unsupervised} to measure the distance between the original image $\mathbf{I}$ and the reconstructed image $\hat{\mathbf{I}}$:
\begin{align}\label{equ-loss-rec}
	\begin{aligned}
		\mathcal{L}_{rec}=-\frac{1}{\vert\Omega\vert}\sum\ln\frac{1}{\sqrt{2}\sigma}\exp-\frac{\sqrt{2}\vert\hat{\mathbf{I}}-\mathbf{I}\vert}{\sigma}\\
		-\frac{1}{\vert\Omega\vert}\sum\ln\frac{1}{\sqrt{2}\sigma}\exp-\frac{\sqrt{2}\vert\hat{\mathbf{I}}_{flip}-\mathbf{I}\vert}{\sigma},
	\end{aligned}
\end{align}
where $\Omega$ is for normalization and $\sigma\in\mathbb{R}^{H\times W}$ is the confidence map estimated by a network to present the symmetric probability of each position in $\mathbf{I}$, $\hat{\mathbf{I}}_{flip}$ is the image reconstructed with the flipped albedo and shape.

\section{More Visualizations}
\begin{figure*}
   \begin{center}
   \includegraphics[width=0.98\linewidth]{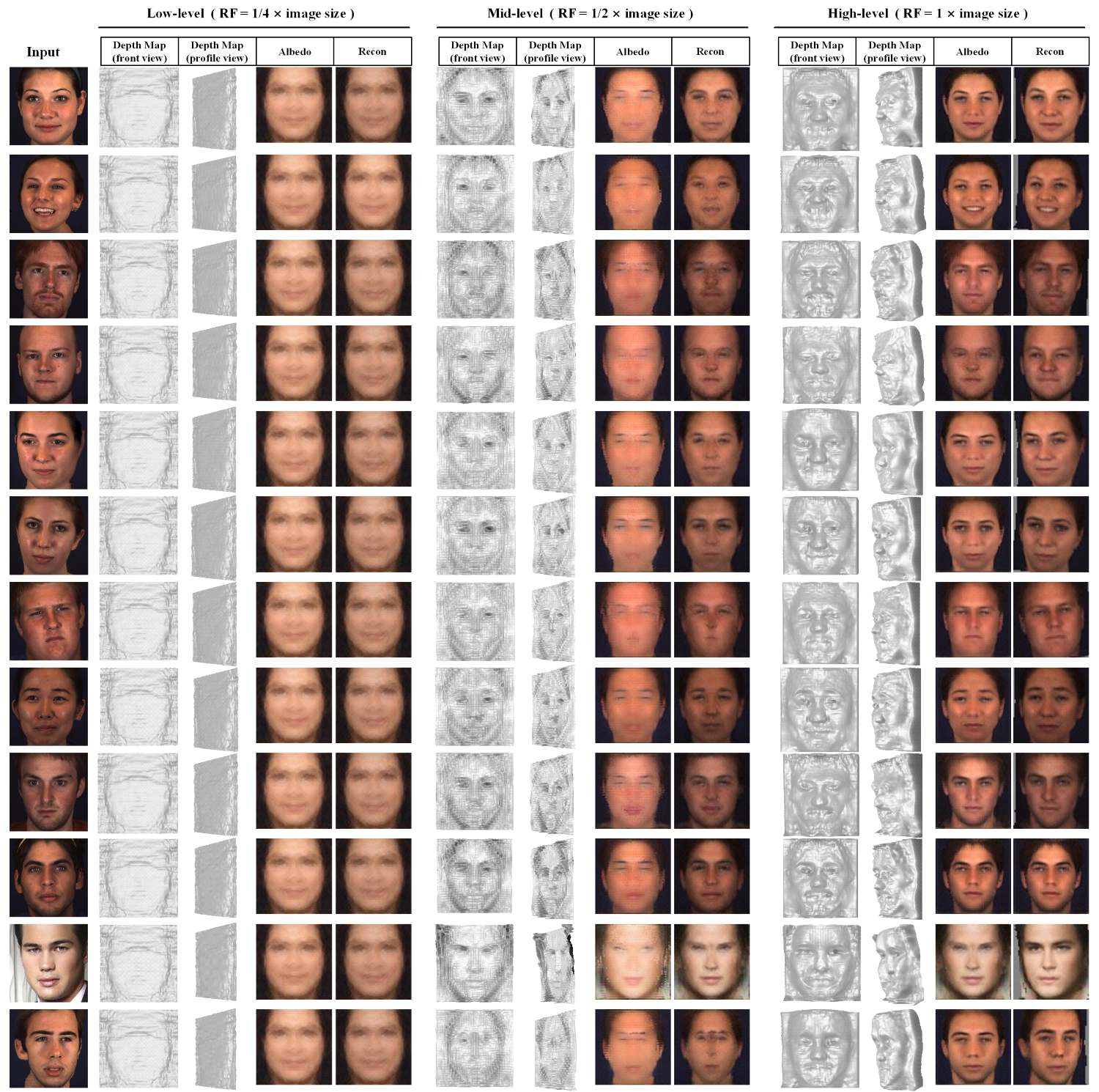}
   \caption{\textbf{Visualization of intermediate representations}.We show the reconstructed depth maps in canonical and profile views, the albedo maps, and the reconstructed results.} 
   \label{fig-supp-vis}
   \end{center}
\end{figure*}

We provide more probe results in Figure \ref{fig-supp-vis}, additionally showcasing the perceived albedo maps along with the reconstructed images. First, it is evident that the perceived geometry evolves from consistently 2D, 2.5D, to 3D as we progress from lower to higher levels, which further substantiates our findings. Secondly, as we ascend to higher levels within the network, there is a marked improvement in the clarity of the perceived albedo, and the reconstructed images increasingly resemble the input images, indicating a decrease in reconstruction loss.  Finally, we observe that reconstruction failures are primarily found at the low and middle levels, please see the squeezed face at the middle level of the last row. 

\section{Geometry Visualization on Different Architecture}
In the main article, we quantitatively analyze the reconstructed depth maps across various neural network architectures, specifically in Section 4.7 titled 'Analysis on More Architectures'.  In this section, we provide visual evidence that complements our quantitative experiments.  The architectures evaluated include VGG16~\cite{simonyan2014very}, ResNet18~\cite{he2016deep}, Swin Transformer (SwinT)~\cite{yu2022degenerate}, and Vision Transformer (ViT)~\cite{dosovitskiy2020image}.
It is worth noting that CNNs and transformers have different methods to generate probe tokens. Unlike transformers that can incorporate additional tokens, CNNs introduce an extra branch at the end of each block. This branch performs average pooling on the final feature map and aggregates it into a single feature. Subsequently, this feature is used to predict depth, albedo, lighting, and view as in graphics probe.

Our findings indicate that these architectures exhibit a consistent progression in depth perception, evolving from 2D, through 2.5D, to 3D representations. Although VGG and ResNet produce low-level depth maps that are not entirely flat, they lack the perception of semantic structures and show uniform geometry across all samples, generating little shading during rendering. Therefore, we also regard it as a 2D representation. This progression supports the generalizability of Marr's theory of vision, suggesting that it applies broadly across different architectures. 

\begin{figure*}
   \begin{center}
   \includegraphics[width=0.99\linewidth]{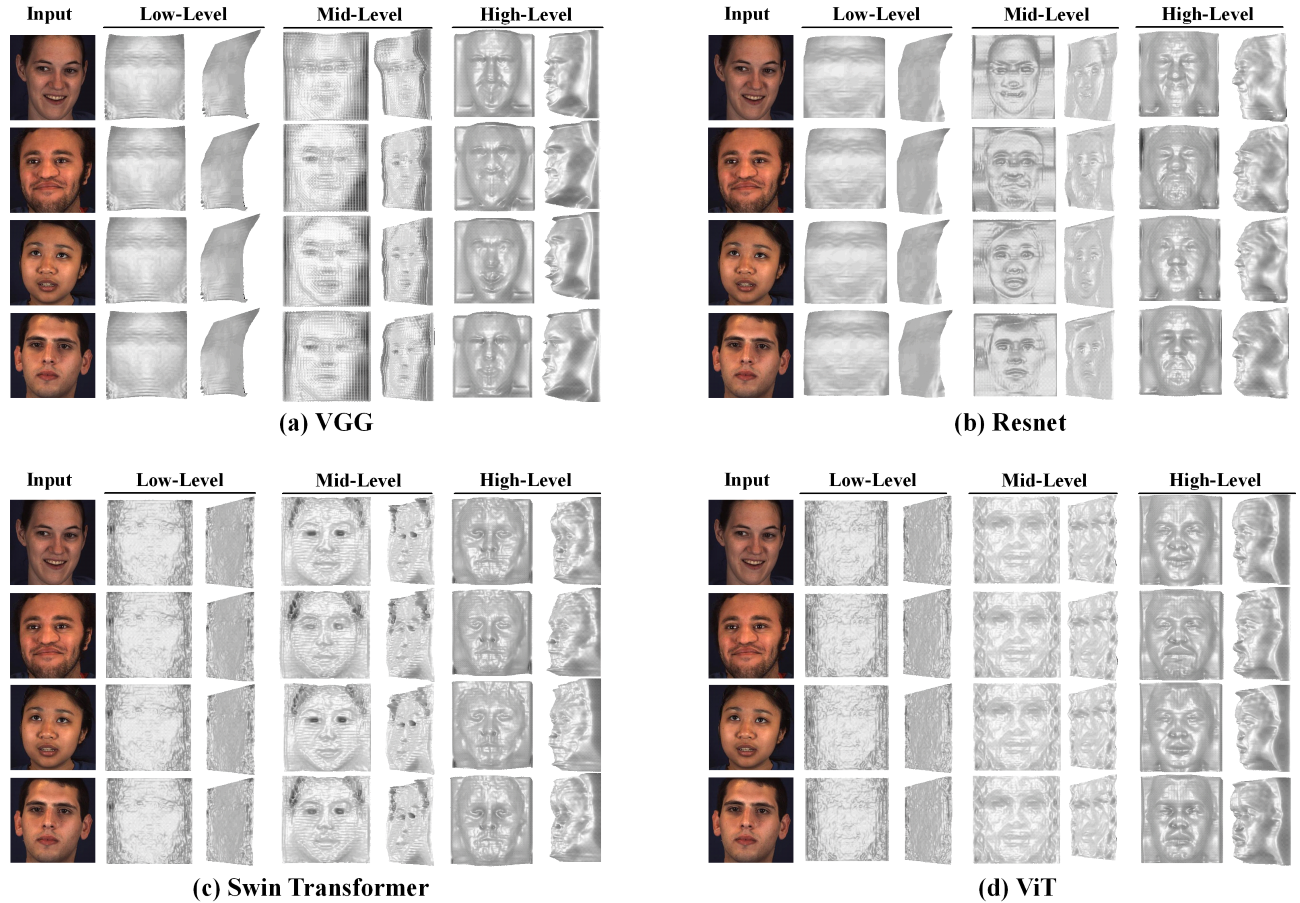}
   \caption{Visualization of intermediate representations on more architectures, including (a) VGG16, (b) Resnet18, (c) Swin-tiny, (d) ViT-tiny. The canonical and profile views of the reconstructed depth maps are demonstrated.} 
   \label{fig-supp-arch}
   \end{center}
\end{figure*}

\section{Visualization of Graphics Probes Trained on Single View}
\begin{figure*}
   \begin{center}
   \includegraphics[width=0.99\linewidth]{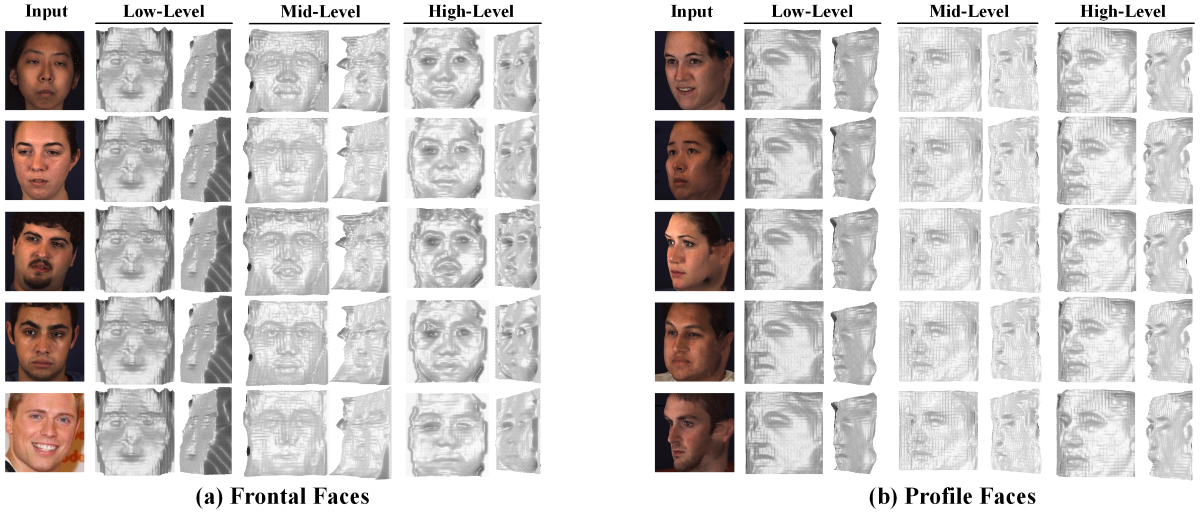}
   \caption{Visualization of the intermediate representations within a network trained on faces in a single viewpoint. (a) The probing results for frontal faces. (b) The probing results for profile faces.} 
   \label{fig-3d-emergence-vis}
   \end{center}
\end{figure*}

In the main article, we investigate the conditions under which a 3D representation can be built. We train a network on a dataset that only contains frontal face images to simulate a scenario where the network has never seen profile views, treating faces as purely 2D objects. In addition to the quantitative analysis presented in the main article, we visualize the intermediate representations of the models in Figure~\ref{fig-3d-emergence-vis}(a). We also perform this experiment using a set of profile faces with the same yaw angle, as shown in Figure~\ref{fig-3d-emergence-vis}(b). In both scenarios, faces can be seen as 2D objects. We observed that the probed geometry dictates the shading of a frontal or profile face, suggesting a 2.5D representation. However, this does not lead to the development of a full 3D model. These findings indicate that if an object is inherently 2D, only a 2.5D representation is formed. It is the observation from diverse viewpoints that facilitates the emergence of a 3D representation.

\small
\bibliographystyle{IEEEtran}
\bibliography{supp-bibliography}